# Multilingual person name recognition and transliteration

**Bruno Pouliquen[1], Ralf Steinberger[1], Camelia Ignat[1], Irina Temnikova[2], Anna Widiger[1], Wajdi Zaghouani[1], Jan Žižka[3]**

| (1) European Commission Joint Research Centre (JRC) | (2) Bulgarian Academy of Sciences | (3) University of Brno |
|---|---|---|
| 21020 Ispra (VA), Italy | Institute for Parallel Processing – Linguistic Modelling Department | Center of Biostatistics and Analyses – Masaryk University, Kamenice 126/3 |
| http://www.jrc.it/langtech | 1113 Sofia, Bulgaria | 602 00 Brno, Czech Republic |
| Name.Surname@jrc.it | Irina.Temnikova@lanzz.org | zizka@cba.muni.cz |

*We present a tool that extracts person names from multilingual news collections and matches name variants referring to the same person. A novel feature is the matching of name variants across languages and writing systems, including names written with the Greek, Cyrillic and Arabic writing system. Due to our highly multilingual setting, we use an internal standard representation for name representation and matching, instead of adopting the traditional bilingual approach to transliteration. This work is part of a news analysis system that clusters an average of 15,000 news articles per day to detect related news within the same and across different languages.*

# Introduction

Many large organisations continuously monitor the media, and especially the news, to stay informed about events of interest, and to find out what the media say about certain persons, organisations, or subjects. Software tools that automatically pre-select the news articles of interest and that pre-process the chosen text collection simplify the daily repetitive task of media monitoring. Crestan & de Loupy (2004) showed that Named Entity extraction and visualisation help users to browse large document collections more quickly and efficiently. This seems plausible as, according to Gey (2000), 30% of the content-bearing words in news are proper names.

In news analysis it is important to know *What* the subject is, *Who* is being talked about, *Where* and *When* things happened, and *How* it was reported. This paper focuses on the occurrence of proper names in news, i.e. the *Who* part of the analysis. Previous work focused on answering the questions *What* (Pouliquen et al. 2004b), *Where* (Pouliquen et al. 2004a) and *When* (Ignat et al. 2003). Due to the highly multilingual work environment in the European Commission – an organisation with twenty official languages – multilinguality of tools and the cross-lingual aspect are of prime importance.

Our analysis is applied to the output of the *Europe Media Monitor* system EMM (Best et al., 2002). EMM is a software toolset that monitors a daily average of 25,000 news articles in currently 30 languages, deriving from 800 different international news sources. For a subset of about 15,000 articles per day in currently eight languages, we apply unsupervised hierarchical clustering techniques to group related articles separately for each language. We then track related news clusters within the same language and across six of the languages (Pouliquen et al. 2004b). The Joint Research Centre's (JRC) name recognition tools are applied to each of these clusters, i.e. each group of related texts is treated as one meta-text, for which person and geographical place names are extracted and keywords are identified. For each cluster, the list of all extracted names is stored in a relational database, together with the lists of the extracted organisations, countries and keywords. Statistics are then built automatically every day to derive knowledge about the co-occurrence of references to named entities in the news. A user interface allows news analysts to search and browse the document collection and the stored meta-data. For more details and for an overview of the process, see Steinberger et al. (2005). The results of the daily clustering and information extraction process can be seen in the JRC's *NewsExplorer* application, which is publicly accessible at http://press.jrc.it/NewsExplorer.

People's names play a special role in the NewsExplorer system. They not only give the reader an overview of who is involved in a certain story. Lists of names are also used to identify news about the same event in different languages and news analysts often search for news regarding a certain person. It is therefore crucially important that spelling variants of the same name are identified as belonging to the same individual. For names identified in languages with different writing systems such as Greek, Russian or Arabic, users furthermore appreciate to see the list of names displayed with Romanised script. The JRC's work on name recognition, name variant identification and name transliteration was thus motivated by the needs of the NewsExplorer users.

After giving some background on name transliteration and referring to related work (Section *Background and related work*), we describe tools to identify names in text (Section *Proper name recognition*) and the mechanism to merge name variants, including those written in Cyrillic, Arabic, and Greek script (Section *Detecting and merging name variants*). This is followed by evaluation results (Section *Evaluation*).

**Table 1: Overview of a recognised person name in nine languages, showing various orthographies for the same person. The words in italics show the recognised trigger word(s).**

| ***Language*** | ***text*** |
|---|---|
| English | …death of *former Prime Minister* **Rafik Hariri**, blamed by many opposition… |
| Spanish | …asesinato del *exprimer ministro* **Rafic al-Hariri**, que la oposición atribuyó… |
| French | …l'assassinat de l'*ex-dirigeant* **Rafic Hariri** et le départ du chef de la … |
| Dutch | na de moord op *oud-premier* **Rafiq al-Hariri** gingen gisteren bijna een… |
| German | *… libanesischen Regierungschef* **Rafik Hariri** vor einem Monat wichtige… |
| Slovene | danjega *libanonskega premiera* **Rafika Haririja**. Libanonska opozicija si… |
| Estonian | möödumisele *ekspeaminister* **Rafik al-Hariri** surma põhjustanud… |
| Arabic | …اغتيال *رئيس الوزراء السابق* **رفيق الحريري** بأياد يهودية وما حدث سابقا … |
| Russian | …*Бывший премьер-министр Ливана* **Рафик Харири**, который… |

# Background and related work

This section gives some background and points to state-of-the-art applications regarding the clustering of news into related articles (See *Grouping of news about the same event or story*), regarding Named Entity Recognition (See *Named Entity Recognition*) and regarding the transliteration of person names and their mapping with European name variants (See *Transliteration of proper names*).

## Grouping of news about the same event or story

As NewsExplorer is fed from hundreds of news sources talking frequently about the same events, the users of the system are faced with a lot of redundant information. Before extracting information from the news, we therefore group the news of one day into clusters of related articles. This has the added advantage that users can easily compare what different sources say about the same event. Grouping the news is also useful to improve the recall of the extracted name information: even if a name was not found in one article, it is likely to be found in another article of the same cluster.

In order to group related news articles into a cluster, we first build a vector representation for each article by extracting a ranked list of keywords and their keyness. *Keyness* is the degree with which the keyword seems to be outstandingly important in the text. For the identification of the keywords and their keyness, we use the statistical log-likelihood test (Dunning, 1993). The reference word frequency list used for this was built using the same text type, i.e. large numbers of news texts. The resulting keyword vector of each document is enhanced by adding additional dimensions representing the countries mentioned in the article. These country dimensions are built by first identifying references to geographical places and then weighting the references with the same log-likelihood formula (see example in Table 2). For details, see Steinberger et al. (2005).

**Table 2: Example of the vector representation of a newspaper article talking about the fact that London was selected over Paris in the competition to host the 2012 Olympic Games. The ranked list of keywords was enhanced by adding country information in ISO format (*gb* stands for United Kingdom, *fr* for France and *es* for Spain) even though the article did not actually mention the country names "France" or "United Kingdom" verbatim.**

| Keyword/Country | Log-likelihood |
|---|---|
| london | 87 |
| *gb* | 25 |
| *fr* | 22 |
| olympic | 16 |
| city | 15 |
| ioc | 14 |
| paris | 11 |
| games | 9 |
| madrid | 9 |
| olympics | 8 |
| east | 5 |
| capital | 5 |
| york | 5 |
| president | 5 |
| *es* | 5 |
| chirac | 4 |
| coe | 4 |
| blair | 4 |
| … | |

In the next step, we use this document vector representation to build hierarchical clustering trees (dendrograms) of all articles of the day, one for each language, using an agglomerative algorithm (Jain et al, 1999). This tree is built by first calculating a pair-wise similarity between all document pairs, using the cosine on their document vectors. The most similar pair of vectors is then combined into a new vector and a new representation is built for this sub-cluster, by merging the keywords and by averaging their keyness. This new node will henceforth be treated like a single document, with the exception that it will have twice the weight of a single document. The hierarchical, binary clustering continues in this way until all documents are part of the cluster. The resulting dendrogram will have clusters of articles that are similar, and a list of keywords and their weight for each cluster. The degree of similarity for each cluster is shown by its intra-cluster similarity value, or cluster cohesiveness.

In a next step, we search the dendrogram for the major news clusters of the day (topic detection), by identifying all sub-clusters with an empirically derived minimum intra-cluster similarity of 50%. For each cluster, we choose the article that is most similar to the cluster's centroid and we use its title as the title for the whole cluster. For further details, see Pouliquen et al. (2004b) and Steinberger et al. (2005).

The idea of grouping news by similar stories is already used by some well known websites like Google news (http://news.google.com/), Altavista news (http://www.altavista.com/news/) and Microsoft Newsbot (http://newsbot.msnbc.msn.com/). The three companies do not explain which techniques they use. However, the main obvious differences between our NewsExplorer and these commercial systems are that we aim at presenting news across languages and that we make use of extracted named entities to browse the news.

# Named Entity Recognition

Though Named Entity Recognition (NER) is a known research area (e.g. MUC-6 1995, Daille & Morin 2000), multilingual Named Entity Recognition is quite new (ACL-MLNER 2003, Poibeau 2003).

People's names can be recognised in text (a) through a lookup procedure if a list of known names exists, (b) by analysing the local lexical context (e.g. 'President' *Name Surname*), (c) because part of a sequence of candidate words is a known name component (e.g. 'John' Surname), or (d) because the sequence of surrounding parts-of-speech (or other) patterns indicates to a tagger that a certain word group is likely to be a name. Sometimes, Machine Learning approaches are used for recognising names within their context by looking at words surrounding known names. For the European languages, it is sufficient to consider only uppercase words as name candidates. Other languages, such as Arabic, do not distinguish case. At the JRC, we currently use methods (a) to (c), but we do not use part-of-speech taggers, because we do not have access to such software for all languages of interest. Until now, the focus has been on people's names, but we also recognise some organisation names.

The JRC's Named Entity Recognition tools differ in some features from other similar tools. These differences are due to the specific needs of the NewsExplorer environment: Firstly, we aim at covering many languages rather than at optimising the tools for a specific language, because NewsExplorer already now covers eight languages and more languages are to come. Secondly, we aim at identifying each name at least once per text or even per cluster of articles and we have no specific interest in identifying every single occurrence of a name in a text. This is due to the fact that users need to see the names mentioned in each *cluster* of news and that they should be able to find all clusters in which a certain name was mentioned. The aim is thus to optimise the recall of the name recognition system *per cluster* instead of per name instance. Thirdly, we aim at recognising names with at least two name parts (typically first name and last name, e.g. ‘German *Gerhard Schröder*’) and ignore references to names consisting of only the last name (e.g. ‘Chancellor *Schröder*’). The decision to ignore one-word name parts is due to the fact that it is safer to use the combination of first and last names to match name variants. The name recognition recall is nevertheless very satisfying because news articles usually mention the full name of a person at least once per article, together with the title or function of the person.

# Transliteration of proper names

Transliteration is the process of representing words from one language using the alphabet or writing system of another language (Arbabi et al., 1994). Transliteration is used to formulate concepts mainly existing in one language (e.g. *Sharia* law) into another, or for reporting about names of people, organisations or places. Transliteration from a language like Arabic would differ depending on the target language. An example is the Arabic name مصطفى, which could be transliterated into English as ‘Mustafa’ or ‘Mustapha’, while a likely French transliteration would be ‘Moustafa’ or ‘Moustapha’[1].

## Specificity of transliterating person names

Many publications, web sites and transliteration schemes exist for languages that use the Cyrillic, Greek, or Arabic alphabets, but most of them apply to general words rather than to person names. The fundamental difference between transliterating natural language words and transliterating names is that the pronunciation of words normally follows some conventions, meaning that hand-crafted linguistic equivalence rules can be used. While the same may be partially true for names of the same language (e.g. Russian names in Russian text), transliteration becomes more difficult when the names found are of international origin – as is often the case in news articles. For instance, in a Russian news article it is likely that names of French, Italian, English or Arabic origin are found. In order to transliterate such international names efficiently, it would be necessary to know the source of the name as this tells us about the target language equivalence. If the origin of the name *Chirac*, for example, is known as being French, then it is pronounced as /ʃiʀak/ and should be transliterated as شيراك in Arabic, or Ширак in Russian. However, if it were an Italian name, it would be pronounced as /kirak/ and transliterated as كيراك in Arabic and Кирак in Russian.

## Dealing with many language pairs

Because of the language dependence of transliteration, previous work in automatic name transliteration has always been carried out for specific language pairs such as Chinese-English or Russian-English, as can be seen in the large enumeration of previous work in Lee et al. (2005). Although it is likely that this limitation to

---

[1] A search on Google gives an idea of the usage of each spelling as:

| | |
|---|---|
| Mustafa | 2,030,000 |
| Mustapha | 709,000 |
| Moustafa | 136,000 |
| Moustapha | 124,000 |

specific language pairs produces better results than our more language-independent approach, such language-dependent approaches are not a useful option in the context of our highly multilingual news analysis system, which aims at dealing with twenty or more languages and where the original language of names is usually not known.

# Transliteration challenges

The transliteration of names from each writing system poses its own challenge. The Cyrillic and Greek scripts seem to be most similar to the Latin script in that they are basically phonetic: letters or groups of letters correspond to specific sounds. The major problems are (a) phoneme-letter equivalences are in an *n-to-n* relationship (i.e. a letter can often be pronounced in different ways and a certain sound can be written with different letters), and (b) the phoneme inventory in different languages (and writing systems) differs: If a language does not know a sound, it will transliterate this sound by another similar one. When back-transliterating the name into the original language, the spelling is thus likely to be wrong. For instance, the German and English sound for the letter ‘h’ is unknown in Russian and is frequently transliterated into ‘Г’, pronounced /g/. Examples are the city name Heidelberg (Гейдельберг’, pronounced /gejdɛlʲberk/) and Harry Potter (Гарри Поттер, pronounced /garipotɛr/). When these names are found in Russian text and are back-transliterated into English or German, they will thus appear as ‘Geidelberg’ and ‘Gari Potter’, or similar.

# Specific challenges for Arabic transliteration

Arabic does not have the sounds /p/, /v/ and /g/. ‘Paul’ is transliterated as بول /bol/, ‘Valery’ as فاليري (/faliry/), and ‘Globe’ as غلوب (/ ʀlo:b/). A name such as ‘Vladimir Putin’ will therefore be transliterated as فلاديمير بوتين (/fladimi:r buti:n/).

Transliteration from Arabic to languages using the Latin alphabet (*romanisation*) is additionally made difficult by the fact that short vowels are usually not written in Arabic. Any romanisation effort therefore typically includes *vowelisation*, i.e. the insertion of the short vowels in the target language (Arbabi, 1994). As Arabic dialects differ in pronunciation, vowelisation is clearly dependent on the dialect. This is presumably the reason why forty different transliterations can be found for the unique spelling of the Arabic name سليمان, including ‘Salayman’, ‘Seleiman’, ‘Solomon’, ‘Suleiman’ and ‘Sylayman’.

# Challenges for languages using ideographs

Transliteration into languages with an ideographic writing system such as Chinese, where each symbol is equivalent to a concept rather than to a sound, has to be tackled in an entirely different way. Chinese has a system of syllables called *Pinyin* (Swofford 2005), a combination of initial and final sounds which can be used to construct about 300 syllables. When transliterating non-Chinese names, a closest syllable-to-syllable approximation is looked up, and for each syllable a Chinese corresponding ideogram can be chosen from the list of different tone variants. The transliteration of an English or German name will thus consist of a concatenation of Chinese syllables. For example, ‘Beethoven’ would be represented in Pinyin as ‘bej-do-fen’.

## Methods for transliterating

Existing automatic name transliteration systems either use hand-crafted linguistic rules, or they use Machine Learning methods (e.g. Lee et al. 2005), or a combination of both. Arbabi et al. (1994), for instance, use linguistic rules and neural networks to vowelise and romanise Arabic names, as well as to filter out unlikely over-generated target word forms. Lee et al. (2005) learn name transliteration from large bilingual Chinese-English lists of proper names, using the Expectation Maximisation algorithm. They do not use pronunciation dictionaries or manually generated phonetic similarity scores. At the JRC, we are using hand-crafted transliteration rules. The output is then processed by further hand-crafted substitution rules in order to produce an *internal standard representation* (see Section *Detecting and merging name variants*).

# Proper name recognition

At the JRC, we add all names detected during our daily news analysis to a database of known names, so that these names can then be recognised in the future by a simple lookup procedure (method (a) described in Section *Named Entity Recognition*). After 18 months of news analysis, the database has grown to about 220,000 distinct names (not counting variants of the same name; see Section *Detecting and merging name variants*). More than 500 new names are inserted every day. For performance reasons, a Unicode (UTF-8) compatible Finite State Automaton is used. A set of regular expressions is generated for each entry of the database as input to the FLEX utility (Paxson 1995), which generates the automaton. In order to exclude the recognition of name variants due to typing errors, the automaton only searches for names that were found at least twice (currently about 68,000, i.e. two thirds of all names have been found only once). To date, the tool thus searches for these 68,000 person names, representing about 84,500 different orthographies.

## Trigger words

To guess new names (method (b) described in Section *Named Entity Recognition*), an extensive list of local patterns was developed in a boot-strapping procedure: We first collected trigger words from various open sources in order to write simple local patterns in PERL to recognise names in a collection of three months of English, French and German news. We then looked at the most frequent left and right hand side contexts of the resulting list of known names. For English alone, we currently have about 1,100 local patterns, consisting of titles (‘Dr.’, ‘Mr’, etc.), country adjectives (such as ‘Estonian’), professions (‘actor’, ‘tennis player’, etc.), specific patterns (such as ‘[0-9]+ year-old’), etc. We refer to these local patterns as *trigger words*. For each added language, native speakers translate the existing pattern lists and use the same bootstrapping procedure to complete the patterns.

Those patterns allow the program to recognise new names (e.g. in “the American doctor John Smith”), but a stored list of such patterns is also useful to give users additional information about persons. In the previous example, for instance, the user will see that *John Smith* probably is an American doctor. When a name is often used with the same trigger words, statistical measures can be used to qualify names automatically. For instance, *George W. Bush* will be recognised as being the American president, *Rafik Hariri* as being the ‘former Lebanese prime Minister’, etc.

Currently the JRC has rules for the following languages: English, French, German, Spanish, Italian and Swedish. To a certain extent we have also some Dutch, Danish, Norwegian, Bulgarian, Estonian and Slovene patterns. A first version of Russian is almost ready, Arabic is under development. The aim is to include all twenty official languages of the European Union and candidate countries.

**Table 3: two examples of patterns used to recognize Tony Blair and Romano Prodi in Slovene texts**

| |
|---|
| Tony(a\|o\|u\|om\|em\|m\|ja\|ju\|jem)?\s+Blair(a\|o\|u\|om\|em\|m\|ju\|jem\|ja)? |
| Roman(a\|o\|u\|om\|em\|m\|ju\|jem\|ja)?\s+Prodi(a\|o\|u\|om\|em\|m\|ju\|jem\|ja)? |

# Dealing with declension

In some languages, especially in Slavonic and Finno-Ugric languages, the proper names are inflected and can have suffixes, as can be seen in the Slovene example 'Tožba proti Donaldu Rumsfeldu zaradi mučenj'. The automaton to recognise names therefore has to allow for a variety of suffixes (in the given example, the suffix 'u' was added to the name *Donald Rumsfeld*). Some of the hand-written rules used at the JRC to detect person and place names simply consist of possible suffix lists for each name. Others are more complex, for instance using substitution functions to detect the Estonian *New Yorgile* as an inflection of *New York* or detecting that the 'o' in *Romano Prodi* is part of the name and should not be identified as the 'o'-suffix in Slovene text. Table 3 shows two sample suffix lists that are required to detect known names in Slovene text. Table 4 lists the rules used to generate automatically inflected variants for Russian names in our database.

**Table 4: Simplified rules to build a pattern recognising all possible declensions of a given name in Russian text.**

<table>
<tr><th>Ending</th><th>Pattern</th><th>Example</th></tr>
<tr><td>-а</td><td>а|ы|и|е|у|ой</td><td>Никит**а** Никит**ы** Никит**е** Никит**у** Никит**ой**<br>*Nikita Nikity Nikite Nikitoy Nikitu*<br>Ольг**а** Ольг**и** Ольг**е** Ольг**у** Ольг**ой**<br>*Oljga Oljgi Oljge Oljgu Oljgoy*</td></tr>
<tr><td>-я</td><td>я|и|ю|е|ей|ёй|ю</td><td>Иль**я** (*Iljya*) Иль**и** Иль**е** Иль**ю** Иль**ёй**<br>Дарь**я** (*Darya*) Дарь**и** Дарь**е** Дарь**ю** Дарь**ей**</td></tr>
<tr><td>-ь</td><td>ь|и|ью|я|ю|ем|е</td><td>Любов**ь** (*Lyubovj*) Любов**и** Любов**ью**<br>Игор**ь** (*Igorj*) Игор**я** Игор**ю** Игор**ем** Игор**е**</td></tr>
<tr><td>-й</td><td>й|я|ю|ем|е|и</td><td>Андре**й** (*Andrey*) Андре**я** Андре**ю** Андре**ем** Андре**е**<br>Анатоли**й** (*Anatoliy*) Анатоли**я** Анатоли**ю** Анатоли**ем** Анатоли**и**</td></tr>
<tr><td>-ел</td><td>ел|ла|лу|лом|ле</td><td>Пав**ел** (*Pavel*) Пав**ла** Пав**лу** Пав**лом** Пав**ле**</td></tr>
<tr><td>-ев</td><td>ев|ьва|ьву|ьвом|ьве</td><td>Л**ев** (*Lev* translated as ‘Leo’) Л**ьва** Л**ьву** Л**ьвом** Л**ьве**</td></tr>
<tr><td>-о<br>-у<br>-е<br>-э<br>-и</td><td>- (not declined)</td><td>Марко (*Marko*)<br>Мари (*Mari*) - *French ‘Marie’*<br>Андрэ (*Andre*)</td></tr>
<tr><td>Default case: consonants</td><td>а|у|ом|ем|е</td><td>Иван (*Ivan*) Иван**а** Иван**у** Иван**ом** Иван**е**<br>Джордж (*Dzhordzh*) Джордж**а** Джордж**у** Джордж**ем** Джордж**е**<br>*English. ‘George’*</td></tr>
</table>

## Storage of names in a database

Names identified in any of the analysed languages are automatically stored in a database, together with information on where and when the name was found and with information on the language of the text. The trigger words found around the name are also stored. Each distinct name is assigned a unique numerical identifier. Variants identified for the same name (see Section *Detecting and merging name variants*) are all stored with the same identifier. To add additional name variants, especially in non-European languages, we automatically search the free Wikipedia[2] online encyclopaedia for all names in our database (see Figure 1). When a Wikipedia entry exists, we add the corresponding URLs to the database to allow users to find additional information about a certain person. Additionally we copy the photograph of the person, when available.

When we detect new names, we use a fuzzy matching tool to automatically detect whether the name is a variant of a name already present in the database (see Section *Fuzzy matching of name variants*).

Table 1 shows the difficulty of recognising names across languages.

---

[2] http://en.wikipedia.org/

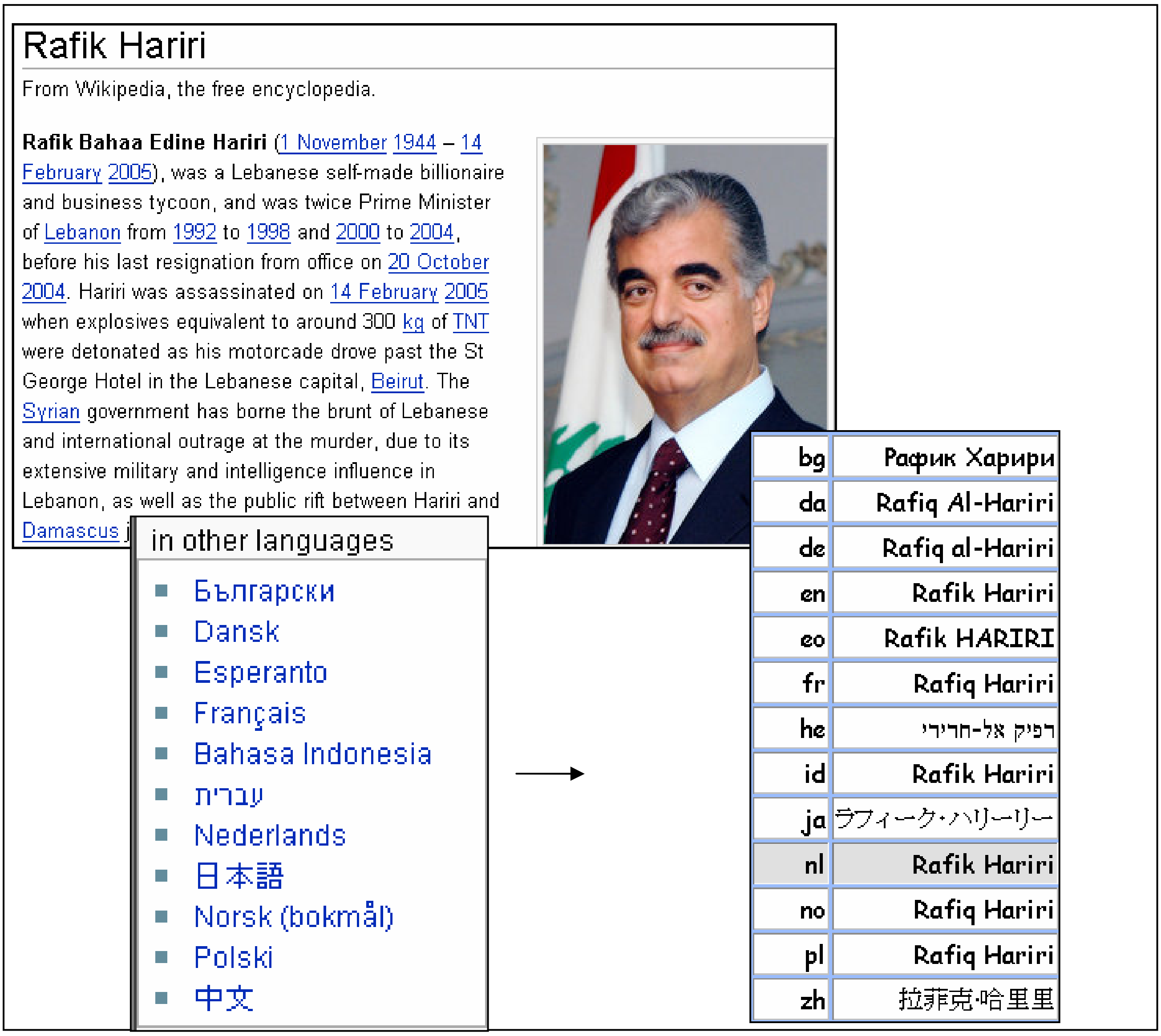

**Figure 1: Entry for Rafik Hariri in the Wikipedia encyclopaedia (http://en.wikipedia.org/wiki/Rafik_Hariri), and some name variants.**

# Detecting and merging name variants

For many person names, several variants are used in the media, not only across languages (see Table 1) but often even within the same language (in 50 English articles published on the 14th April 2005, we found four orthographies: *Rafik Hariri*, *Rafik al-Hariri*, *Rafiq Hariri* and *Rafiq al-Hariri*). In order to allow users find information about certain persons independently of the name spelling, we aim at storing all name variants together with one unique numerical identifier.

Using the similarity of name orthography (described in Section *Fuzzy matching of name variants*), we currently merge name variant candidates automatically if they are found in the same news cluster and if their similarity score is high enough (70% similarity or higher). As clusters can consist of between 2 and 100 articles talking about the same event (for details see Pouliquen et al. 2004b), it is quite likely that two variants of the same person name are found in the same cluster.

As the system to match names *across* languages is still under development, cross-lingual name variant merging is currently done only if two variants are extremely close (i.e. similarity more than 95%). When a new name is detected, its similarity with all other names is computed. Then the process automatically merges similar names (see Table 5 for examples compiled for one day). For the others (similarity between 80% and 95%), the system displays a list of new names similar to previous ones (variant candidates), asking for a human confirmation before merging them. As shown in the examples in Table 6, all names for that day need to be merged. Even the case of *Daniel Cicarelli* turned out to be a typo so that the two names should be merged[3].

**Table 5: List of extremely similar names found in the news of a single day (30 May 2005). These variants are automatically merged.**

[3] The article did in fact intend to talk about Daniella Cicarelli ('reciente separación de la modelo brasilena Daniel Cicarelli'); last accessed on 1/06/2005 at http://www.lanacion.com.ar/deportiva/nota.asp?nota_id=708643&origen=rss.

| ***New name*** | ***Merged with existing name:*** |
|---|---|
| Abdüllatif Sener | Abdullatif Sener |
| Abubakar Tanko | Aboubakar Tanko |
| Allan McDonald | Alan McDonald |
| Bahiya al-Hariri | Bahia al-Hariri |
| Brian Herta | Bryan Herta |
| Eid Cabalu | Eid Kabalu |
| Hassan Mohamed Nur | Hassan Mohamed Nuur |
| Ismail al-Hadithi | Ismail al Hadithi |
| Johana Melka | Johanna Melka |
| José Luis Lingeri | Jose Luis Lingeri |
| Luis Fernández | Luis Fernandez |
| Michael Haefrati | Michael Haephrati |
| Mohamed Dhia | Mohammed Dhiaa |
| Nikolas Sarkozy | Nicolas Sarkozy |
| Salomé Zurabishvili | Salome Zurabishvili |
| Sergei Brin | Sergey Brin |
| Stanley Fisher | Stanley Fischer |
| Surat Ikramov | Sourat Ikramov |
| Trudi Stevenson | Trudy Stevenson |
| Werner Schneyder | Werner Schneider |

As we do not currently consider the context of names, it could happen that the system merges names such as ‘Mariana Gonzalez’ (a Venezuelan fencer) and ‘Mariano Gonzalez’ (an Argentinean football player). The system therefore allows manual intervention to correct incorrectly merged names or to merge two variants that have not been detected automatically.

As shown in Table 5, Table 6 and Footnote 3, quite a few misspelled names appear in the news, but it is important to capture them anyway in order to improve the *Recall*.

**Table 6: List of rather similar new names (30 May 2005). Before merging these variants, manual confirmation is required.**

| | |
|---|---|
| Russ Young | Ross Young |
| Gary Shafer | Gary Sheffer |
| Mohammed Dhia | Mohammad Dhiya |
| Brian Vilora | Brian Viloria |
| Saad al-Harir | Saad al-Hariri |
| Pierre Gadonnaix | Pierre Gadonneix |
| Abudullahi Yusuf | Abdullahi Yusuf |
| *… (altogether 24 propositions) …* | |
| Daniel Cicarelli | Daniella Cicarelli |

Due to the usage of different scripts in Greek, Russian and Arabic, the merging of names in these languages partially differs from the process used for languages written with the Roman alphabet.

## Normalisation of the name orthography

Name variants across languages are often due to the omission of diacritics. For example, a British newspaper can sometimes refer to *François Mitterrand* as *Francois Mitterrand*. A number of further regular variations we observed are the singling of double consonants, transliterations of *f* by *ph* (e.g. *Ralph Schumacher*), alternative usage of *w* or *v* in Russian names (e.g. *Wladimir* vs. *Vladimir*), alternative spellings of the sound 'u' as *u* or *ou*, etc. In languages such as Lithuanian, transliterations are common (e.g. *Buš* for *Bush*). We therefore decided to develop an *internal standard representation*, ISR, which has the pragmatic aim of linking the variants, without wanting to make theoretical claims of any sort.

Before calculating a similarity between pairs of names found in texts of different languages, all names are *standardised* using a set of approximately 30 substitution rules. Examples are:

accented character → non-accented equivalent
double consonant → single consonant
ou → u
wl (beginning of name) → vl
ow, ew (end of name) → ov, ev
ck → k
ph → f
ž → j
š → sh

This list of substitution rules can also contain the most frequent exceptions not covered by the generic rules (e.g.: Джеймс => 'James' to avoid the basic transliteration as 'geyms'). Examples of names after this standardisation are:

Otto Schily → Oto Shili
Wladimir Ustinow → Vladimir Ustinov
Vladimir Oustinov → Vladimir Ustinov
Abdalah Džburi → Abdalah Djburi
Abdallah Joubouri → Abdalah Juburi
Malik Saïdoullaïev → Malik Saidulaiev
Malik Saidullajew → Malik Saidulajev

## Transliteration of non-Latin scripts

For Greek, Russian and Arabic, which do not use the Latin script, we use hand-written transliteration and adaptation rules to represent names with the Latin alphabet. Transliteration consists of a number of substitution rules that replace one or more non-Latin characters by one or more Latin characters. For Greek, for instance, the following substitutions apply:

λ → l
θ → th
μπ → b

After the transliteration, the normalisation rules described in the previous section (*Normalisation of the name orthography*) are applied. The results of the transliteration and standardisation are often phonetic (e.g. 'Bil Klinton', 'Jak Shirak', etc.), but they are similar enough to the standard representation to produce good results in the fuzzy matching process (see Section *Fuzzy matching of name variants*). Example results for Greek, Cyrillic and Arabic transformations are:

Κόφι Ανάν (Greek) → Kofi Anan
Кофи Аннан (Russian) → Kofi Anan
Кофи Анан (Bulgarian) → Kofi Anan
كوفي عنان (Arabic) → Kufi Anan
कोफी अन्नान (Hindi) → Kofi Anan

At the JRC, we have developed transliteration rules for the following writing systems: Greek, Cyrillic (Russian, Bulgarian and Ukrainian), Arabic (including Farsi and Urdu) and Devanagari (Hindi and Nepali)[4].

## Fuzzy matching of name variants

In order to identify potential name variants (like those in Table 6) we carry out a pair-wise comparison of all transliterated and standardised names. If the similarity of the pair of names is above a certain threshold, the names are variant candidates.

[4] Writing the rules to transliterate a new language can be rather quick. Rules for the Devanagari script, for instance, took about 2 hours.

For the similarity calculation we combine three similarity measures. We currently take the average of the three measures, but we plan to evaluate the relative impact of each of them in order to optimise their relative weight for the similarity calculation.

The three measures are based on letter ngram similarity: we calculate the *cosine* of the letter ngram frequency lists for both names, separately for bigrams and for trigrams; the third measure is the *cosine* of bigrams based on strings without vowels. We do not use phonetic transliterations of names as these are reported to be less useful than *string-like* approaches (Zobel & Dart, 1995). Moreover, phonetic transliteration rules are different from language to language (e.g. *Chirac* would in Italian be pronounced as /kirak/) and finding the transliteration rules for many languages would be difficult.

Figure 2 gives an overview of the process for comparing a French name with its Russian counterpart written with Cyrillic letters.

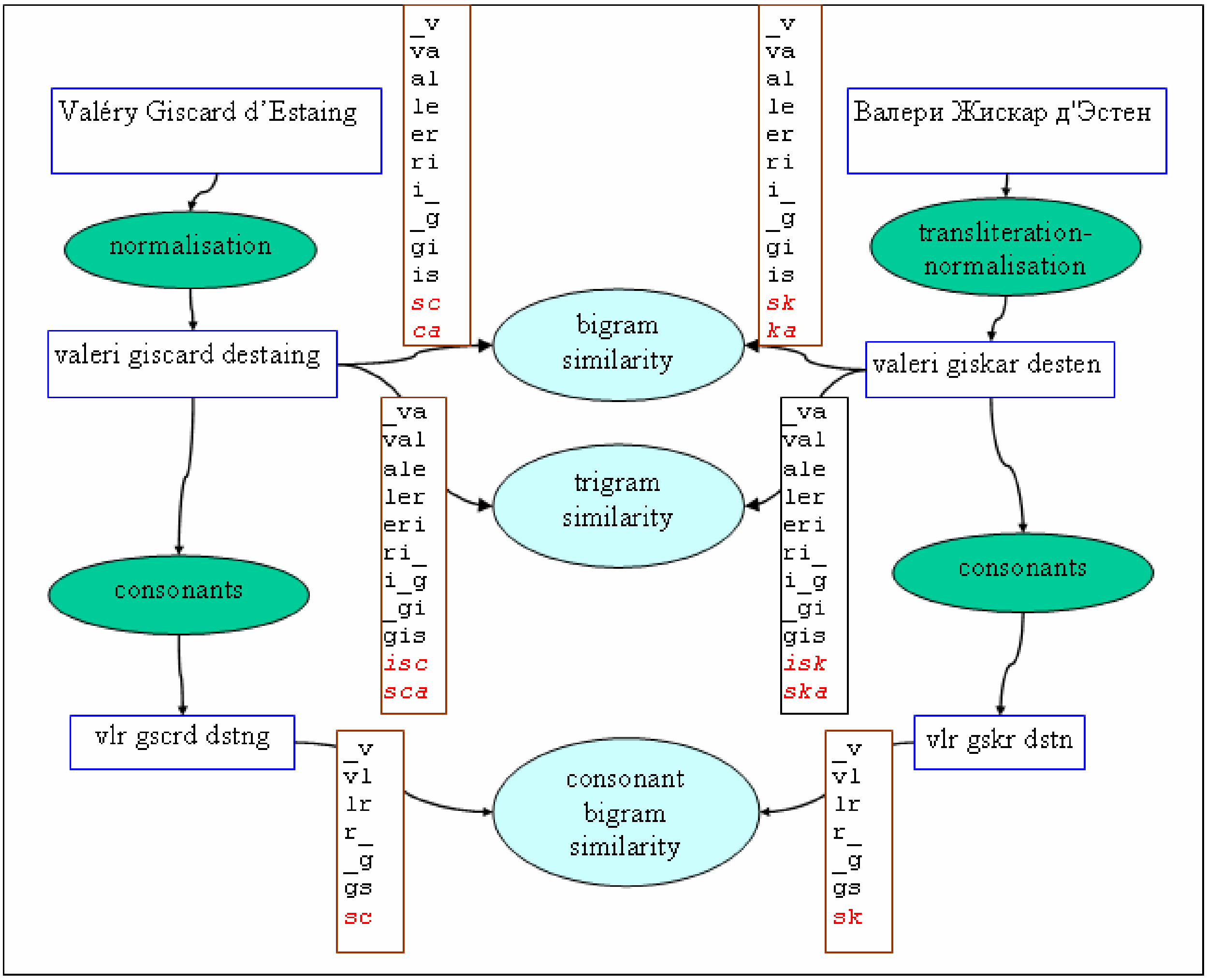

**Figure 2: Example of the matching process between a Cyrillic (Russian) and a Latin (French) name.**

## Special variation to deal with Arabic

Standard Arabic writes long vowels and often omits short ones. When comparing names written in Arabic with names written with the Latin alphabet, we therefore delete vowels from the latter before computing the similarity. For instance, the internal standard representation for the name *Condoleezza Rice* is ‘kondoleza rice’. The same name written in Arabic is كوندوليزا رايس. The result of the transliteration and standardisation of the Arabic version of the name is ‘konduliza rais’. The *cosine* of bigrams between these two representations without vowels (‘kndlz Rc’ and ‘kndlz Rs’) is rather high (0.875) so that the two names written with the Arabic and the Latin scripts are successfully identified as name variants.

Figure 3 summarises the matching process for an Arabic name.

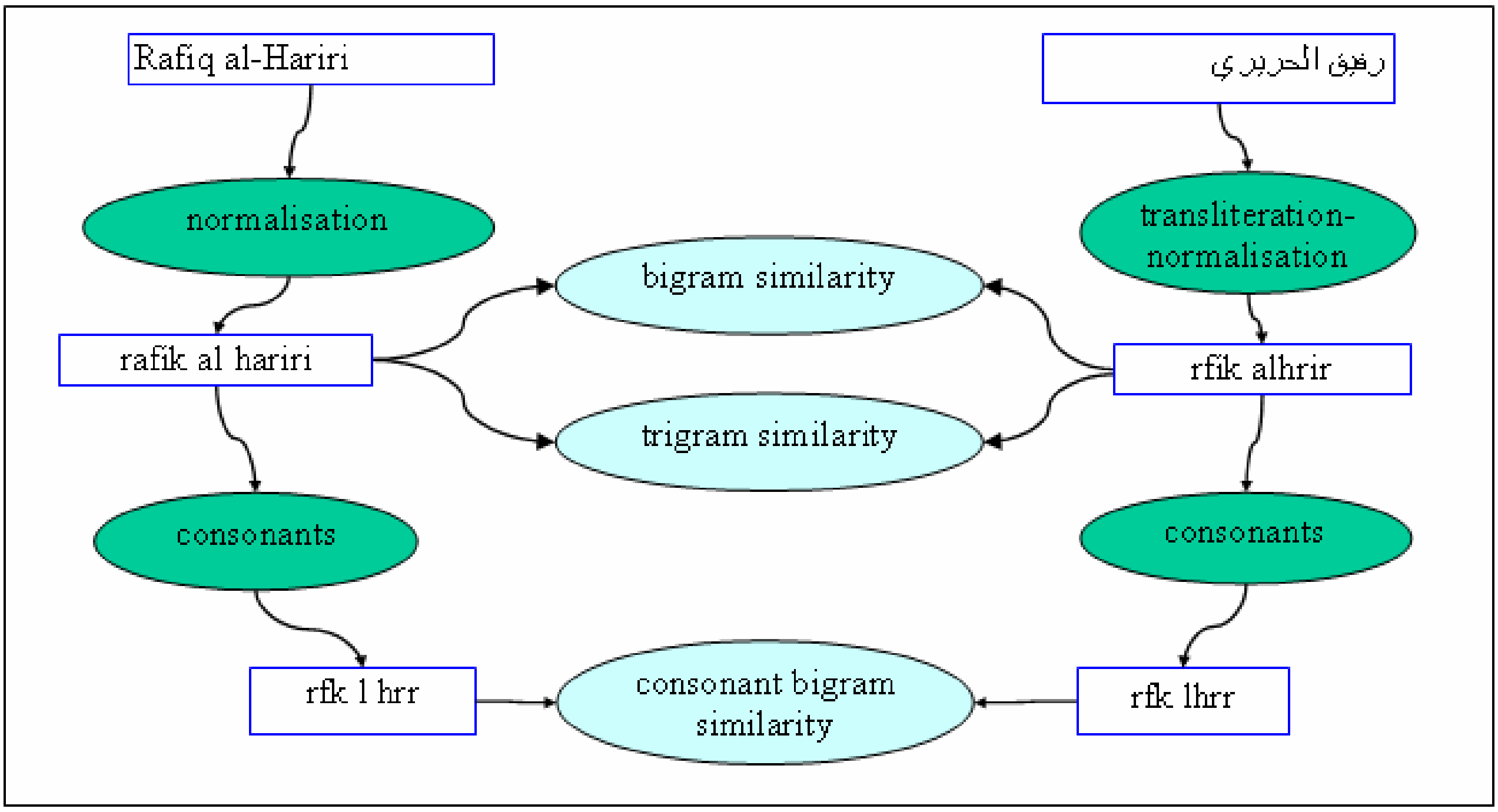

**Figure 3: Arabic/Latin name matching example.**

# Evaluation

## Evaluation of name recognition

Our focus is not on optimising Named Entity Recognition for one language, but rather on finding an approach that is easily and quickly adapted to new languages. We nevertheless launched an evaluation on the performance of the tool for various languages:

In each language we chose a random selection of about 100 newspaper articles. We applied our person name recognition tool. Experts listed all person names that were present in the text. For each article we then compared if each of the automatically recognised person names was also selected by the expert (to get *Precision*), and if all the manually extracted names were also automatically found (to get *Recall*). We combine those two values using the F-measure[5]**.**

We have to emphasise that, unlike in traditional name recognition evaluation, our aim was to identify the presence or non-presence of a name in the text, and that it was not our concern to identify each and every mention of the name. Furthermore we restricted our evaluation to the recognition of person names, ignoring organisations and toponyms. The results are summarised in Table 7.

**Table 7: Evaluation of person name recognition in various languages.** ***The number of rules (i.e. trigger words) gives an idea of the expected coverage for this language. The third and fourth columns show the size of the test set (number of texts, number of manually identified person names).***

| ***Language*** | ***# rules*** | ***# texts*** | ***# names*** | ***Average Precision*** | ***Average Recall*** | ***Average F-measure*** |
|---|---|---|---|---|---|---|
| ***English*** | 1100 | 100 | 405 | 92 | 84 | 88 |
| ***French*** | 1050 | 103 | 329 | 96 | 95 | 95 |
| ***German*** | 2400 | 100 | 327 | 90 | 96 | 93 |
| ***Spanish*** | 580 | 94 | 274 | 85 | 84 | 84 |
| ***Italian*** | 440 | 100 | 298 | 92 | 90 | 91 |
| ***Russian*** | 447 | 61 | 157 | 81 | 69 | 74 |

The results are less good than for Named Entity Recognition systems that use part-of-speech taggers. Those are optimised for a given language, and do not aim at such high multilinguality. Our *Precision* is nevertheless reasonably high. In our setting, where we try to detect names in *clusters* of news instead of in individual articles, the lower *Recall* is not a big problem because names are usually found in at least one of the articles so that the person information for the cluster is often complete.

---

5 $$Fmeasure = 2\frac{precision \cdot recall}{precision + recall}$$

The low Recall score could be due to the nature of our heterogeneous test set: The set not only includes articles from many different domains (politics, sports results, discussions of television programmes, etc.), but also from international newspapers from all over the world (especially for the English language texts). The system has to analyse articles such as: 'Phe Naimahawan, of Chiang Mai's Mae Ai district, has been selected (…) to represent Thailand in a swimming event (…). Phe is being helped by Wanthanee Rungruangspakul, a law lecturer'. Without part-of-speech tagging, it is difficult to guess that 'Phe Naimahawan' is a person name. However, in the same text, we were able to guess the name 'Wanthanee Rungruangspakul' thanks to the trigger word 'law lecturer'.

The lower *Precision* for German was predictable as in German every noun is uppercased, which often results in the system recognising common nouns as proper names. In the example: "Die österreichische Eishockey Nationalmannschaft bekommt während der Heim-WM noch Verstärkung", 'Eishockey Nationalmannschaft' (*Ice hockey national team*) is wrongly triggered as being a person name by the trigger word 'österreichische' (*Austrian*).

The relatively bad scores for Spanish are due to various facts. One of them was that we did not have any Basque first names in our name lists and that many Basque names were found in the test set (ex: Gorka Aztiria Echeverría**)**. Another reason was that our system frequently only recognised part of the typical Spanish compound names (ex: Elías Antonio Saca, where the process recognised only Antonio Saca). Finally, several organisation names were wrongly classified by the algorithm as person names.

The explanation for the lower Russian results mainly is that our name database contained only a dozen Russian names so that the system had to guess most names, which is harder than looking up known names.

## Evaluation of transliteration

An unbiased evaluation of the variant matching algorithm for names written with the Latin script is not possible because all frequent variants are already stored in the database, and some of them had already been manually checked or were added via the Wikipedia search (see Section *Storage of names in a database*). We would only be able to test the system on new names, but for these we would not find variants in the database. Testing the system on previously unseen variants is not particularly useful either. Instead, we evaluated how accurately the system identified the Latin equivalent of names written with Cyrillic (Russian) and Arabic letters (Arabic and Farsi languages). For this purpose, three native-speakers prepared a short list of randomly selected names that they found in the news of the day. We then verified whether or not the system proposed the European version of this name as the most similar (with a minimum similarity threshold of 50%). We must highlight that each of the names was compared to the orthographies of 68,000 other persons.

This test allows us to see whether the transliteration, standardisation and fuzzy matching tool works properly. Moreover, it allows us to validate whether our database contains the most important names.

### Russian name variant identification

Out of 53 names analysed, only one did not have a Latin equivalent in the database, but the system still returned a wrong proposal. In two other cases, the wrong person was identified. The remaining 50 names were successfully matched to the correct person. Both *Precision* and *Recall* were thus 94.3%.

The person not present in our database was *Robert Bradtke* (the American Secretary for European and Eurasian Affairs). This name was wrongly matched with *Robert Bartko* (a German cyclist).

The two false hits were *Jean-Claude Juncker* (transliterated as ‘Jan-Klod Yunker’ and matched with *Carsten Jancker*), and *Heinz Fischer* (transliterated as ‘Khaynts Fisher’ and matched with *Joschka Fischer*).

## Arabic name variant identification

All of the 30 names selected had a Latin-script equivalent in the database. However, two names were not found and three names were assigned to the wrong person. The remaining 25 names were matched successfully. *Precision* is thus 89.3% and *Recall* is 83.3%.

Among the good examples, *Jean-Pierre Raffarin*, transliterated as ‘Jan-Biar Rafaran’, was still matched; and similarly *Arnold Schwarzenegger*, transliterated as ‘Arnuld Shuarznijr’. Even short names such as *Jack Straw*, transliterated as ‘Jak Stru’, were found.

The two names not found were due to bad transliteration: *John Garang* has the Arabic variant جون قرنق which was transliterated as ‘Jon Qrnq’ and was not similar to any names in the database. The same is true for جورج كلوني, which was transliterated as ‘Jurj Kloni’ and should have been identified as *George Clooney*.

Wrongly matched names were *John McCain*, transliterated as ‘Jon Mak Kin’ (and matched with *Jean Makoun*), *Colin Powell* transliterated as ‘Kuln Baul’, and *Michael Jackson* as ‘Maikl Jakson’. An obvious solution would be to manually add transliteration rules for the most common names (George, John, Michael, etc.).

## Farsi name variant identification

A native speaker selected 22 names from online articles on BBC world service[6]. All of them were actually in our database, 20 were found as being the most similar, but the system did not find two names. Both *Precision* and *Recall* were thus 90.9%.

Among the good examples, مانوئل باروسو (Manuel Barroso) was transliterated as ‘Manuil Barusu’ and was successfully matched with the president of the EU commission. Even پرویز مشرف (Pervez Musharraf), transliterated as ‘Pruiz Mshrf’, was still matched.

The two errors were due to wrong transliteration. پرویز مشرف (Jean-Claude Juncker) was transliterated as ‘Zan Klud Ionkr’ and was not found. داگلاس وود (Douglas Wood) was transliterated as ‘Daglas Uud’ and wrongly matched with Douglas Hurd, Douglas Wood ranked second but with a very low similarity of 0.32.

---

[6] http://www.bbc.co.uk/worldservice/

# Conclusion and future work

We have developed multilingual software tools that analyse large news collections, cluster related news into groups, extract named entities from the clusters, and merge name variants belonging to the same person. The automatically generated news meta-data is stored in a relational database that allows users to explore the news collection, by searching for full text or for meta-data features. Through the daily news analysis in currently eight languages, the fully automatic system learns more co-occurrence relations between the different entity types every day. Most of the functionality is publicly available on the JRC's *NewsExplorer* web site (see example in Figure 4; http://press.jrc.it/NewsExplorer/).

While many of the tools mentioned in this paper are already integrated in NewsExplorer, others still have to mature and stabilise. The cross-lingual matching of name variants already produces useful results for an interactive system, but the merging of name variants cannot yet be fully automated because it still produces errors. The name mapping across writing systems is not being used at all at this moment because NewsExplorer is currently only working for languages using the Roman script. We would like to explore how the cluster context of two names and the lists of their stored trigger words can be used to improve the quality of the name merging tool. Comparison of time series like in Shinyama & Sekine (2004) could improve the *Precision*.

We also plan to dedicate more time to improve the name recognition itself. We have started working on recognising organisation names, but there is much more to do. We would also like to explore systems to automatically (or semi-automatically) extend patterns to new languages.

Currently we use the content of the Wikipedia entries only to get cross-lingual links and the photograph of the person. Interesting research would be to mine these Wikipedia texts automatically for further information. The relationship between people, for instance, could be confirmed if a given person is mentioned in somebody else's page.

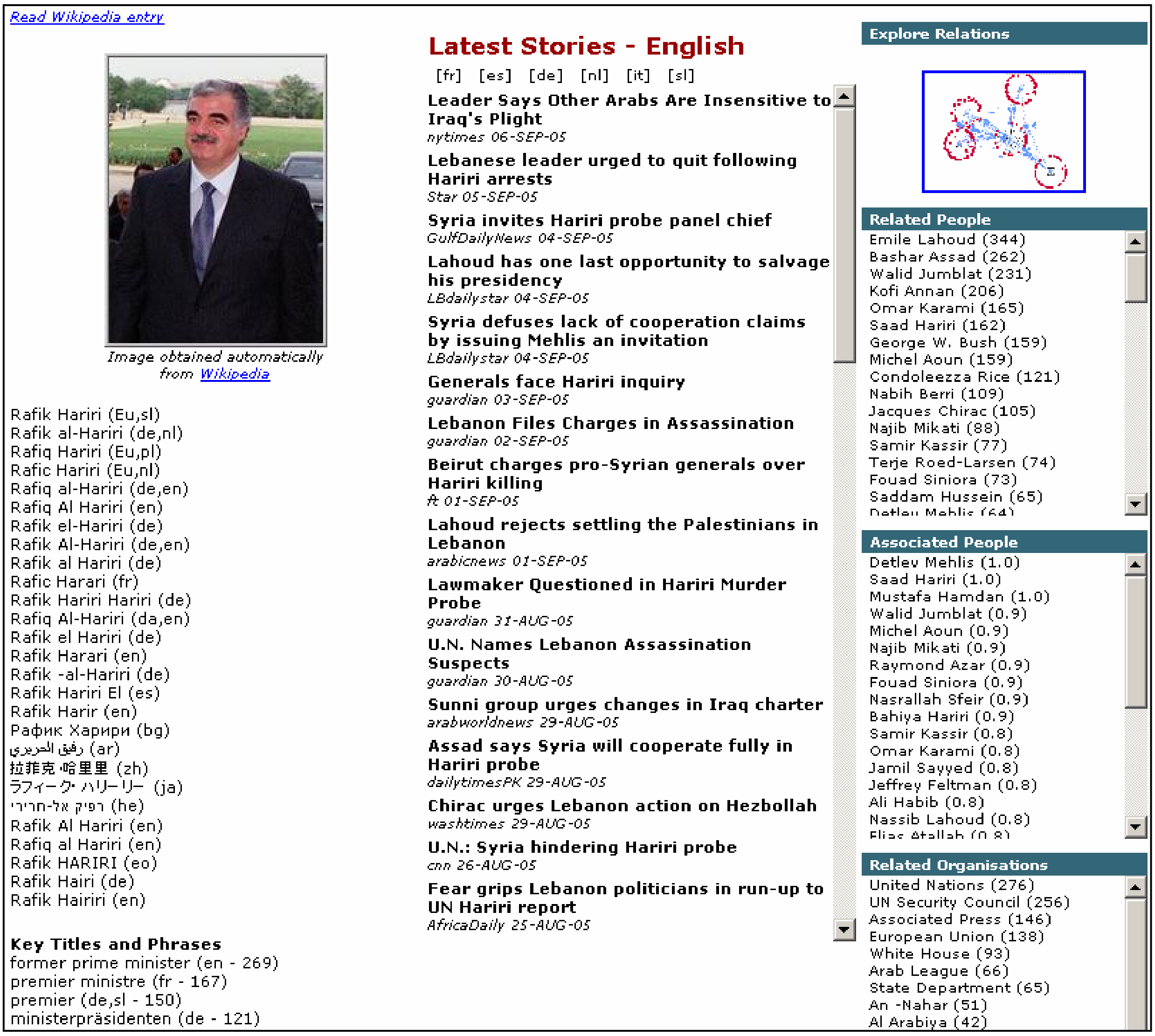

**Figure 4: NewsExplorer entry for Rafiq Hariri**

# Acknowledgements

We thank the whole team of the Web Technology sector at the JRC for providing us with the highly valuable news data to test the tools, as well as for their technical support. We also want to thank Carlo Ferigato (JRC) who introduced us to various fuzzy matching techniques. We thank Tomaž Erjavec from the Jožef Stefan Institute in Ljubljana and Ann-Charlotte Forslund (JRC) for helping us with the Slovene and Swedish languages, and Helen Salak (JRC) for providing us with knowledge about Farsi.